\documentclass[10pt]{article} 
\usepackage[accepted]{rlc}

\usepackage{amssymb}            
\usepackage{mathtools}          
\usepackage{mathrsfs}           
\usepackage{graphicx}           
\usepackage{subcaption}         
\usepackage[space]{grffile}     
\usepackage{url}                

\usepackage{cleveref}
\crefname{figure}{Figure}{Figure}
\crefname{table}{Table}{Table}
\crefname{algorithm}{Algorithm}{Algorithm}
\crefname{equation}{Eq.}{Eq.}
\usepackage{adjustbox}
\usepackage{listings}

\usepackage{multirow}
\usepackage[dash,dot]{dashundergaps}

\usepackage{xcolor}
\definecolor{mycolor}{rgb}{0.9,0.9,0.9}
\newcommand\pythonstyle{\lstset{
language=Python,
basicstyle=\ttfamily\small,
commentstyle=\color{gray},
numberstyle=\tiny\color{gray},
stringstyle=\color{red},
frame=none,
showstringspaces=false,
keywords={},                      
keywords=[2]{def,return},
keywordstyle=[2]\color{blue},
keywords=[3]{gumbel_loss,expanded_gumbel_loss},
keywordstyle=[3]\color{red},
backgroundcolor=\color{mycolor}
}}

\lstnewenvironment{python}[1][]
{
\pythonstyle
\lstset{#1}
}
{}

\newcommand\pythoninline[1]{{\pythonstyle\lstinline!#1!}}

\title{Stabilizing Extreme Q-learning by Maclaurin\\ Expansion}

\author{Motoki Omura  \\
    omura@mi.t.u-tokyo.ac.jp \\
    The University of Tokyo
    \And
    Takayuki Osa  \\
    osa@mi.t.u-tokyo.ac.jp \\
    The University of Tokyo \\
    RIKEN
    \And
    Yusuke Mukuta  \\
    mukuta@mi.t.u-tokyo.ac.jp \\
    The University of Tokyo \\
    RIKEN
    \And
    Tatsuya Harada  \\
    harada@mi.t.u-tokyo.ac.jp \\
    The University of Tokyo \\
    RIKEN
    }
    
\begin{document}

\maketitle

\begin{abstract}
In offline reinforcement learning, in-sample learning methods have been widely used to prevent performance degradation caused by evaluating out-of-distribution actions from the dataset.
Extreme Q-learning (XQL) employs a loss function based on the assumption that Bellman error follows a Gumbel distribution, enabling it to model the soft optimal value function in an in-sample manner. It has demonstrated strong performance in both offline and online reinforcement learning settings.
However, issues remain, such as the instability caused by the exponential term in the loss function and the risk of the error distribution deviating from the Gumbel distribution.
Therefore, we propose Maclaurin Expanded Extreme Q-learning to enhance stability. In this method, applying
Maclaurin expansion to the loss function in XQL enhances stability against large errors. 
This approach involves adjusting the modeled value function between the value function under the behavior policy and the soft optimal value function, thus achieving a trade-off between stability and optimality depending on the order of expansion.
It also enables adjustment of the error distribution assumption from a normal distribution to a Gumbel distribution.
Our method significantly stabilizes learning in online RL tasks from DM Control, where XQL was previously unstable. Additionally, it improves performance in several offline RL tasks from D4RL.

\end{abstract}

\section{Introduction}
Deep reinforcement learning has demonstrated good performance in many tasks, including robotics \citep{schulman2017proximal, haarnoja2018soft} and games \citep{mnih2013dqn, mnih2015dqn, silver2016alphago}.
During learning, the goal is to acquire the optimal policy by learning a value function, and the learning of the value function involves the Bellman update.
Recently, \cite{garg2023extreme} proposed that based on the Extreme Value Theorem, the Bellman error follows a Gumbel distribution rather than the normal distribution assumed by traditional least squares methods.
Consequently, they proposed an algorithm called Extreme Q-learning (XQL), which employs Gumbel Regression, a maximum likelihood estimation assuming a Gumbel error distribution.
This method demonstrated excellent performance, primarily in offline RL.
However, a significant issue with Gumbel Regression is its instability. The loss function contains an exponential term that can lead to too large or too small gradients, resulting in divergence or slow convergence.
As a remedy, they used stabilization measures such as clipping and the max-normalization trick, but it was still unstable, particularly in online RL.
Another problem is that the error distribution may not exactly match the Gumbel distribution.
\cite{garg2023extreme} demonstrated that based on the i.i.d. among state-action pairs and time steps, it becomes a Gumbel distribution, but in reality, the independence is not guaranteed due to the use of the same neural network for all states and actions.
Additionally, in the actual Bellman updates of algorithms, many elements such as entropy maximization, target networks, and Clipped Double Q-learning are incorporated, which also affect the error distribution.
As suggested in \cite{garg2023extreme}, the resulting distribution may resemble a mix of normal and Gumbel distributions.

\begin{figure}[t]
  \centering
  \begin{tabular}{cc}
    \includegraphics[width=0.48\columnwidth]{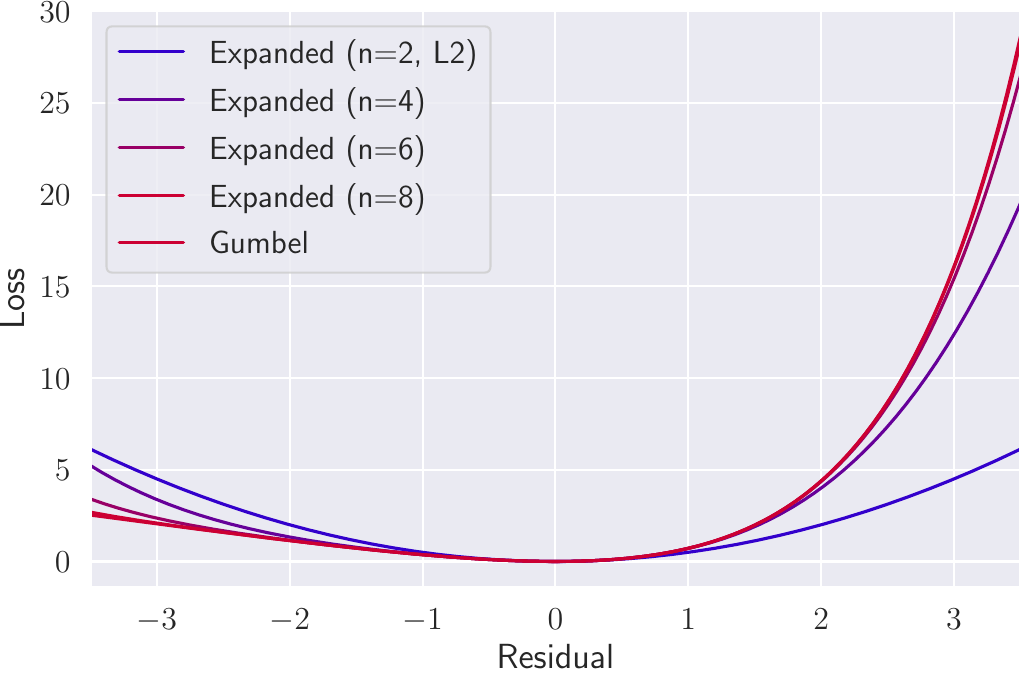} & 
    \includegraphics[width=0.48\columnwidth]{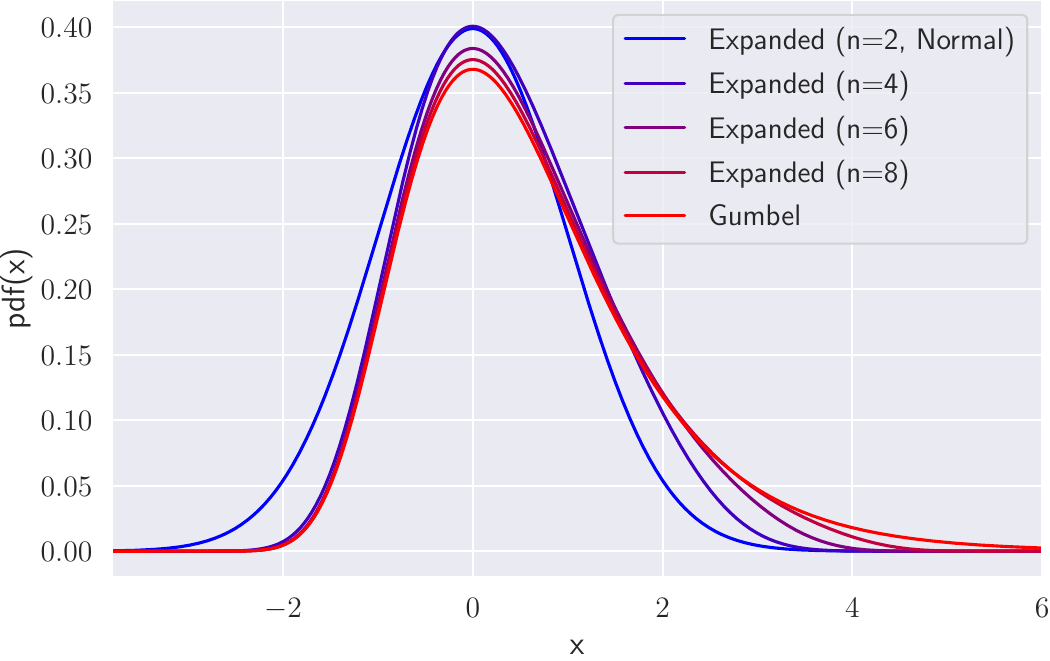} 
  \end{tabular}
  \caption{\textbf{Left}: Illustration of Gumbel loss ($\beta=1$) alongside Expanded Gumbel loss derived from it. When $n=2$, the loss aligns identically with the L2 loss. \textbf{Right}: Depiction of the error distribution assumed by Gumbel loss and Expanded Gumbel loss. Specifically, when $n=2$ (corresponding to L2 loss), a normal distribution is assumed.}
  \label{fig:egloss}
\end{figure}

Thus, in this study, we propose a both simple and practical algorithm, Maclaurin Expanded Extreme Q-learning (MXQL), which not only stabilizes the Gumbel loss but also allows for the adjustment of the error distribution assumption from normal to Gumbel.
The proposed method uses Expanded Gumbel loss, which is a Maclaurin expansion of the Gumbel loss.
By reducing the order of expansion $n$, this loss, as shown to the left of \cref{fig:egloss}, mitigates excessively large or small gradients compared to the Gumbel loss.

As $n$ increases, the loss function converges to the Gumbel loss, while for $n=2$, it becomes the L2 loss. With the Gumbel loss, the estimated value function becomes a soft optimal value, and with the L2 loss, it becomes the value function under the behavior policy. These learning methods correspond to soft Q-learning \citep{a2017softq} and SARSA, respectively. By adjusting $n$ between $2$ and $\infty$, the method of estimating the value function can be adjusted between soft Q-learning and SARSA. In other words, a larger $n$ leads to an unstable but optimal value function estimation, while a smaller $n$ results in a stable but non-maximizing estimation, offering a trade-off between stability and optimality. This is analogous to adjusting the parameter $\tau$ in IQL's expectile loss from $1$ to a smaller value for stabilization. IQL adjusts between SARSA and Q-learning using the parameter $\tau$, while MXQL adjusts between SARSA and soft Q-learning using the parameter $n$.

The assumed error distribution follows a normal distribution when $n=2$ because the loss is the L2 loss. When $n$ is large, it follows a Gumbel distribution. Therefore, by adjusting $n$ between $2$ and a larger value, the assumed error distribution can be adjusted between the normal and Gumbel distributions. In practice, the Bellman error is influenced by various factors such as the target network and double Q-learning, which suggests that, according to the central limit theorem, it may approximate a normal distribution.
The assumed error distribution is illustrated to the right of \cref{fig:egloss}, showing that as $n$ increases, the distribution transitions from normal to Gumbel.

In the experiments, similar to \cite{garg2023extreme}, we compared performance using DM Control tasks \citep{tassa2018dmc2,tunyasuvunakool2020dmc1} for online RL and D4RL tasks \citep{fu2020d4rl} for offline RL.
In online RL scenarios, where XQL was previously unstable, we observed improved stability. Furthermore, in offline RL, our method demonstrated superior performance compared to existing methods, including XQL, across several tasks.

The contributions of this study are as follows:
\begin{itemize}
    \item Verifying XQL's instability through preliminary experiments and demonstrating its instability when the data distribution significantly deviates from the assumed error distribution.
    \item Proposing a novel loss function, the Expanded Gumbel loss, by applying a Maclaurin expansion to the Gumbel loss in XQL, aimed at stabilizing XQL.
    \item Demonstrating improved stability and performance by using Maclaurin Expanded Extreme Q-learning, which incorporates the Expanded Gumbel loss into XQL, across numerous tasks in both online RL and offline RL.
\end{itemize}

\section{Background}

\subsection{Reinforcement Learning}
We address the reinforcement learning (RL) problem within the framework of a Markov decision process (MDP) represented by $(\mathcal{S},\mathcal{A},\mathcal{P},r,\gamma,d)$.
Here, $\mathcal{S}$ represents the state space, where $s$ denotes a specific state. $\mathcal{A}$ represents the action space, with $a$ denoting a specific action. $\mathcal{P}(s_{t+1} | s_t, a_t)$ defines the transition probability from one state to another given an action. $r(s,a)$ is the reward function, representing the reward received after taking action $a$ in state $s$. $\gamma$ represents the discount rate. Lastly, $d(s_0)$ represents the initial state's probability density.
The policy $\pi(a | s)$ is defined as the probability of taking a specific action given the state.
The objective in RL is to discover the policy that maximizes the expected sum of discounted rewards, denoted as $\mathbb{E}[R_0 | \pi]$. The return $R_t$ is calculated as $R_t = \sum^{T}_{k=t} \gamma^{k-t} r(s_k, a_k)$, representing the total discounted rewards from time $t$.

\subsection{Extreme Q-learning}
In this section, we explain XQL, which forms the foundation of this research.
First, based on the Gumbel Error Model, we explain how the Bellman error follows the Gumbel distribution.
Then, we outline the Gumbel Regression for learning the value function, effective for addressing the error distribution of the Gumbel distribution.
Subsequently, we examine the role of the hyperparameter $\beta$ and discuss the inherent limitations of XQL.

\subsubsection{Gumbel Error Model}

The Gumbel Error Model suggests that the Bellman error follows a Gumbel distribution.
Given multiple instances of $\hat{Q}$, each representing a different simulation of the model, the target defined by the Bellman optimal operator is given as:

\begin{equation}
    \begin{aligned}
    \hat{\mathcal{B}}^\ast \hat{Q}_t & = r + \gamma \max_{a'}\hat{Q}_t(s', a') = r + \gamma \max_{a'}(\bar{Q}_t(s', a') + \epsilon_t(s', a')) 
    \end{aligned}
\end{equation}

Here, $\hat{Q}_{t}(s,a) = \bar{Q}_t(s,a) + \epsilon_t(s,a)$. The error $\epsilon_t$, an i.i.d. variable with zero mean, is combined with the mean value $\bar{Q}_t$. Thus, the mean value $\bar{Q}_{t+1}$ for the estimated value $\hat{Q}_{t+1}$ is given by:

\begin{equation}
    \begin{aligned}
    & \bar{Q}_{t+1}(s, a) = r + \gamma \mathbb{E}_{s'|s, a}[\mathbb{E}_{\epsilon_t}[ \max_{a'}(\bar{Q}_t(s', a') + \epsilon_t(s',a'))]]
    \end{aligned}
\end{equation}


Despite $\epsilon_t$ having a zero mean, the maximization operation introduces a positive expectation bias into $\bar{Q}_{t+1}$, resulting in an overestimation bias in Q-learning, specifically when using function approximators.
Furthermore, if $\epsilon_t$ is uncorrelated with $s_t$, $a_t$, and across different time steps, according to the Extreme Value Theorem, $\epsilon_t$ will converge to a Gumbel distribution over time. The error distribution violates the least squares method's assumption that the error follows a normal distribution.

\subsubsection{Gumbel Regression}
Based on the properties of the Gumbel Error Model, they proposed Gumbel Regression, a maximum likelihood estimation assuming a Gumbel distribution as the error distribution. The loss function in Gumbel Regression is as follows:

\begin{equation}
\label{eq:gloss_xh}
L(h) = \mathbb{E}_{x_i \sim \mathcal{D}} [e^{(x_i - h)/\beta} - (x_i - h)/\beta -1]
\end{equation}

where $x_i$ is the samples and $h$ is the parameter to be estimated. 
By minimizing this loss function, the estimated parameters $h$ are found as $h = \beta \log \mathbb{E}_{x_i \sim \mathcal{D}}[e^{x_i/\beta}]$. This expression is analogous to the log-partition function (LogSumExp), where the summation is replaced by an expectation.


In XQL, this property is applied to Q-learning, directly estimating the soft-value function ($V^{*}(s) = \beta \log \sum_a \mu(a \mid s) \text{exp}(Q(s, a)/\beta)$) in maximum entropy RL. The loss function becomes:

\begin{equation}
\label{ep:gloss}
L(V) = \mathbb{E}_{s,a \sim \mu} [e^{(Q(s,a) - V(s))/\beta} - (Q(s,a) - V(s))/\beta - 1]
\end{equation}

where $\mu$ is the behavior policy that generated the sampled state-action pairs. As the estimation of the value function eliminates the need for samples from the current policy, it avoids the computation of Q-values using out-of-distribution actions, thereby enhancing performance, particularly in offline RL scenarios.
In practical online RL applications, the last policy is used for $\mu$, and a trust region update \citep{a2015trpo} is performed.

The Q-function is learned using the least squares method as follows:

\begin{equation}
\label{ep:loss_q}
L(Q) = \mathbb{E}_{s,a, s' \sim \mathcal{D}} [(r(s, a) + \gamma V(s') - Q(s,a))^2]
\end{equation}






\subsubsection{Role of $\beta$}
The soft optimal value estimated in \cref{ep:gloss} is derived from the following objective function in Maximum Entropy RL.

\begin{equation}
   J(\pi) = \mathbb{E}_{s, a \sim \pi}\left[Q(s,a)\right]- \beta D_{\text{KL}}(\pi || \mu) \\
\end{equation}

Here, $\beta D_{\text{KL}}(\pi || \mu)$ serves as a conservative term against deviations from the behavior policy, with $\beta$ modulating the level of conservatism. As can be seen from \cref{ep:gloss}, this parameter, $\beta$, also impacts learning stability: small $\beta$ values can increase gradients, risking divergence, while large $\beta$ values may reduce gradient magnitudes, decelerating convergence. Therefore, while $\beta$ modifies conservatism, it does not ensure stability across all values.

\subsubsection{Limitations}
The first limitation concerns instability, which complicates the adjustment of conservatism through the parameter $\beta$.
Techniques such as clipping and max normalization were employed to achieve stabilization:

\begin{python}
def gumbel_loss(pred, label, beta, clip):
    z = (label - pred)/beta
    z = torch.clamp(z, -clip, clip)
    max_z = torch.max(z)
    max_z = torch.where(max_z < -1.0, torch.tensor(-1.0), max_z)
    max_z = max_z.detach()
    loss = torch.exp(z - max_z) - z*torch.exp(-max_z) - torch.exp(-max_z)
    return loss.mean()
\end{python}

Nevertheless, instability persists, and in certain tasks, learning can collapse, necessitating a restart \citep{garg2023extreme}.
This instability is demonstrated in the following section through preliminary experiments.

The second limitation is that the error distribution may not be a Gumbel distribution. 
The Gumbel Error Model assumes independence between state-action pairs and steps; however, maintaining this independence becomes challenging when the same network is utilized for all training samples.
Moreover, practical algorithms integrate techniques such as target networks and Clipped Double Q-learning, which additionally influence the error distribution.
Plots in the appendix of \cite{garg2023extreme} demonstrate distributions that deviate from the Gumbel distribution.
They also suggest that the error distribution might combine normal and Gumbel elements. We hypothesize that this mixed distribution, influenced by various factors and the Central Limit Theorem, resembles an intermediate between normal and Gumbel distributions.


\section{Maclaurin Expanded Extreme Q-learning}
As mentioned above, XQL has limitations; it can be unstable, and its assumptions regarding the error distribution may not be met.
Therefore, we propose a method named Maclaurin Expanded Extreme Q-learning (MXQL), which stabilizes XQL and facilitates adjustment between normal and Gumbel distributions in the error distribution assumption.


\subsection{Instability related to $\beta$}
In Extreme Q-learning, $\beta$ was adjusted for conservatism, learning efficiency, and stability.
In other words, even if one aims to adjust the level of conservatism using $\beta$, certain $\beta$ values may prove unstable and unusable depending on the data.
To illustrate this, we conducted a simple preliminary experiment.

\begin{figure}[t]
  \centering
  \setlength{\tabcolsep}{2pt}
  \begin{tabular}{ccc}
    \includegraphics[width=0.31\columnwidth]{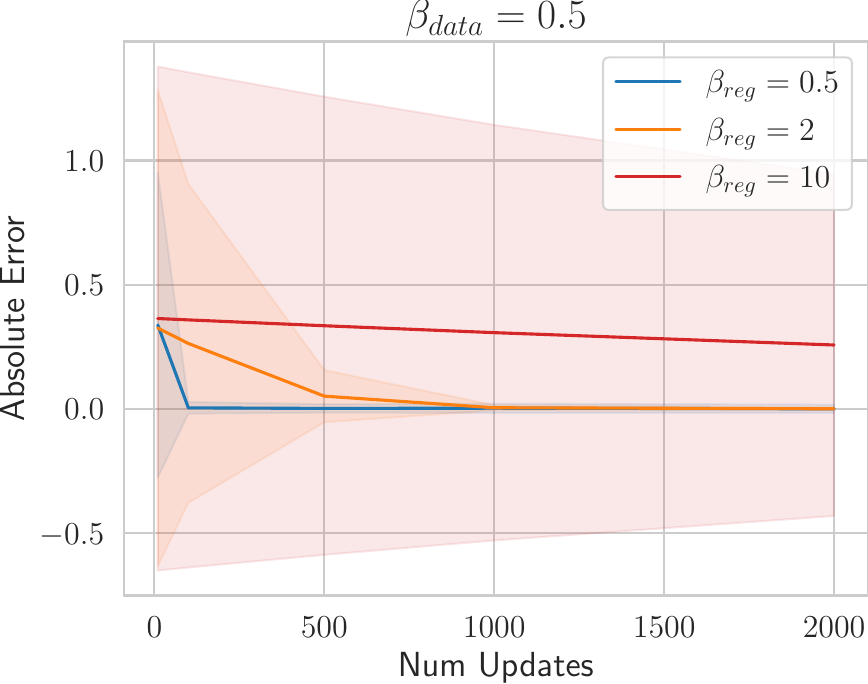} &
    \includegraphics[width=0.31\columnwidth]{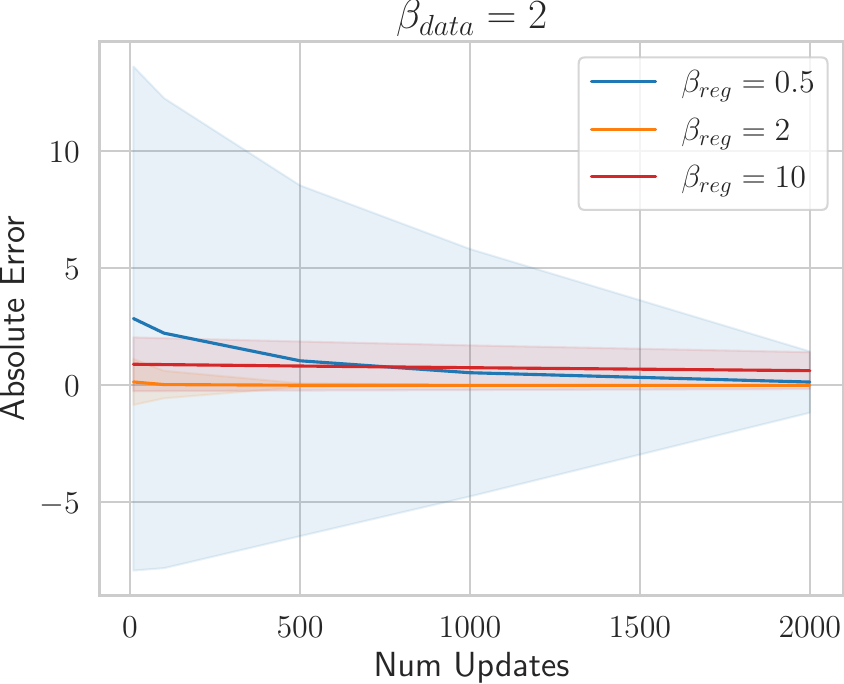} &
    \includegraphics[width=0.33\columnwidth]{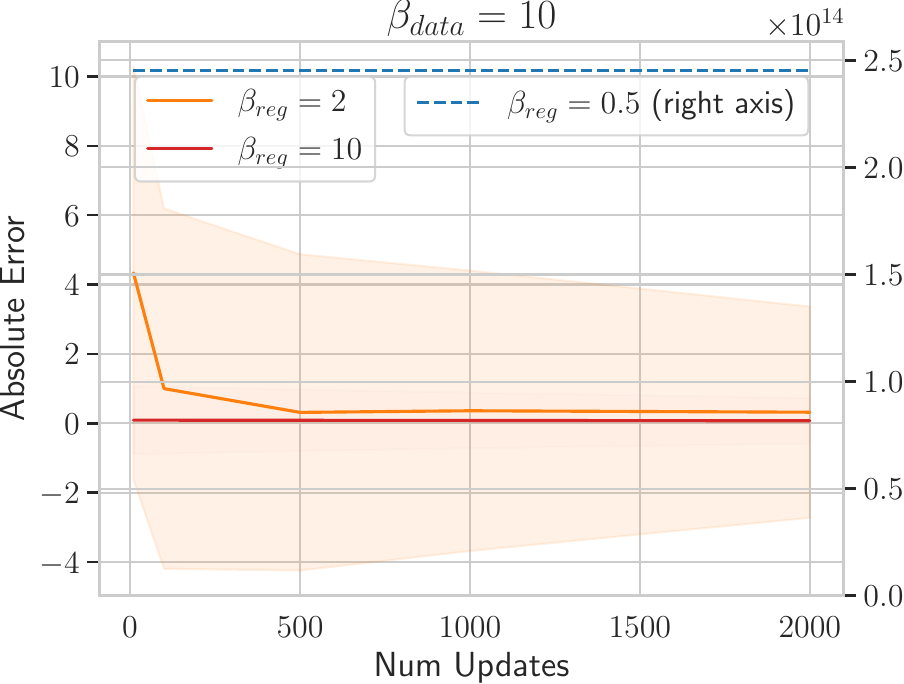} 
  \end{tabular}
  \caption{Results of Gumbel Regression with varying parameters $\beta_{data}$ and $\beta_{reg}$ set to (0.5, 2, 10). The absolute error refers to the absolute difference between the estimated parameter $h$ and the true value, defined as $\log \sum e^{x_i/\beta_{reg}}$. This error was measured at update counts of [10, 100, 500, 1000 2000]. The experiment was conducted 100 times to obtain average values, and the standard deviation is depicted as shaded areas.}
  \label{fig:regression}
\end{figure}

For scalar data represented as $x_i$, parameter $h$, and Gumbel noise $\epsilon_i$, we perform Gumbel Regression using \cref{eq:gloss_xh} as $x_i = h + \epsilon_i$ where $\epsilon_i \sim - \mathcal{G}(0, \beta_{reg}), x_i \sim - \mathcal{G}(0, \beta_{data})$. This implies that when $\beta_{reg}$ and $\beta_{data}$ are equal, the assumptions concerning the error distribution in Gumbel Regression are fully met.
We varied the $\beta$ of the data ($\beta_{data}$) and the $\beta$ used in Gumbel Regression ($\beta_{reg}$) respectively, estimated them, and analyzed how differences between $\beta_{data}$ and $\beta_{reg}$ affected the estimation results. 
The estimated $h$ was assessed based on its deviation from the true value ($\log \sum e^{x_i/\beta_{reg}}$).
The results are presented in \cref{fig:regression}.

When $\beta_{data}$ and $\beta_{reg}$ are equal, that is, when the distribution of the data assumed in Gumbel Regression is the same as the distribution of the data used for estimation, the accurate estimation can be achieved with fewer updates.
When $\beta_{data}$ and $\beta_{reg}$ are different, the required number of updates increases. Furthermore, if $\beta_{reg}$ is too small relative to $\beta_{data}$, the gradient diverges and learning collapses.
Similar outcomes have been noted in higher-dimensional and online RL tasks, demonstrated by \cite{garg2023extreme}, with learning collapse requiring restarts in specific scenarios.

These results suggest that in Gumbel Regression, it is necessary to choose a $\beta$ that approximates the distribution of the learning data closely, thereby narrowing the range within which conservativity can be adjusted.
Therefore, we propose a method to stabilize Gumbel Regression adaptable to diverse $\beta$ values.

\subsection{Expanded Gumbel Loss}
We introduce the Expanded Gumbel loss, which applies a Taylor (Maclaurin) expansion to the Gumbel loss at the point where the residual ($x_i - h$) equals zero.

\begin{equation}
L(h) = \mathbb{E}_{x_i \sim \mathcal{D}} \left[ \sum_{j=2}^{n}  \frac{(x_i - h)^j}{j! \beta^j} \right]
\end{equation}

where $n$ is the order of expansion. As shown on the left side of Figure \ref{fig:egloss}, the Expanded Gumbel loss exhibits a less steep gradient on the right side and a steeper gradient on the left side compared to the Gumbel loss. 
This modification resolves issues such as the collapse of learning and slow convergence observed in the preliminary experiments.
Moreover, by adjusting the expansion order $n$, the degree of stabilization can be altered: smaller values of $n$ lead to greater deviation from the Gumbel loss but enhance stability.
It should be noted that $n$ is selected from even numbers to ensure the loss remains positive.
In MXQL, similar to XQL, the Expanded Gumbel loss is employed for estimating the value function, and the loss function is defined as follows:

\begin{equation}
\label{eq:mxql}
L(V) = \mathbb{E}_{s,a \sim \mu} \left[ \sum_{j=2}^{n}  \frac{(Q(s,a) - V(s))^j}{j! \beta^j} \right]
\end{equation}

Learning the Q-function uses \cref{ep:loss_q} in the same manner as in XQL.

As shown in \cref{eq:mxql}, when $n=2$, the loss function becomes the L2 loss.
The V-function represents the expected value of Q for actions taken under the behavior policy.
In other words, the learning of the value function follows a SARSA-based approach, and the estimated value function is the value under the behavior policy rather than the optimal policy.
When $n$ approaches $\infty$, \cref{eq:mxql} transforms into the Gumbel loss. The value function estimated by the Gumbel loss corresponds to the soft optimal value in maximum entropy RL. Thus, the learning of the value function follows a soft Q-learning-based approach. 
This indicates a trade-off between stability and optimality: when $n$ is large, the estimated value function is more optimal but less stable, whereas when $n$ is small, the value function is more stable but but does not fully maximize.

This trade-off is similar to the parameter $\tau$ in IQL \citep{kostrikov2022iql}, which adjusts the expectile loss. In IQL, when $\tau = 0.5$, the loss becomes the L2 loss, corresponding to $n=2$ in MXQL, while $\tau = 1$ in IQL corresponds to $n = \infty$ in MXQL. IQL adjusts between SARSA and Q-learning using the parameter $\tau$, while MXQL adjusts between SARSA and soft Q-learning using the parameter $n$. Since IQL typically uses values of $\tau$ less than $1$, such as $0.7$ or $0.9$, this suggests the necessity of stabilizing the loss in XQL towards the L2 loss.

In the following section, we present the analysis of the error distribution from the perspective of the Expanded Gumbel loss.


\subsection{Perspective of Error Distribution}
The Gumbel loss was derived from the maximum likelihood estimation that assumes a Gumbel distribution for the error distribution.
Next, we consider the error distribution assumed by the proposed Expanded Gumbel loss.
The error distribution for each $n$ is shown on the right side of Figure \ref{fig:egloss}.
When $n$ is large, the Expanded Gumbel loss approaches the Gumbel loss due to the nature of the Taylor expansion, and consequently, the assumed error distribution also approaches the Gumbel distribution.
When $n=2$, which is the smallest value, the Expanded Gumbel loss takes a form equivalent to L2 loss.
This loss corresponds to the least squares method, which assumes a normal distribution for the error distribution.
As $n$ increases, the error distribution shifts from a normal to a Gumbel distribution.
In other words, by adjusting the expansion order, it is possible to modulate the error distribution between a normal and a Gumbel distribution.
It is conceivable that error distributions influenced by various factors tend to have properties close to a normal distribution due to the Central Limit Theorem.
Moreover, preliminary experiments confirmed the importance of the closeness between the assumed error distribution and the data distribution, suggesting that adjusting the distribution using $n$ is likely to be effective.

We name our method, which substitutes XQL's Gumbel loss with Expanded Gumbel loss, Maclaurin Expanded Extreme Q-learning (MXQL).
Our approach is based on XQL, simply changing the loss function for V-function. The implementation details are provided in the appendix.
In MXQL, the expansion order $n$ is a hyperparameter. However, it should be noted that the clipping size is also a hyperparameter in XQL, and the number of hyperparameters has not changed compared to XQL.


\section{Experiments}
In our experiments, we assessed the stability and performance of Maclaurin Expanded Extreme Q-learning (MXQL) in comparison with Extreme Q-learning (XQL) across both online RL and offline RL scenarios for various $\beta$ values.
The implementation is based on the official XQL framework, and we adhered to the same hyperparameters as those used in the XQL setup.
Further details are provided in the appendix.

\subsection{Online RL}
\begin{figure}[t]
  \centering
  \setlength{\tabcolsep}{2pt}
  \begin{tabular}{lcr}
    \includegraphics[width=0.3\columnwidth]{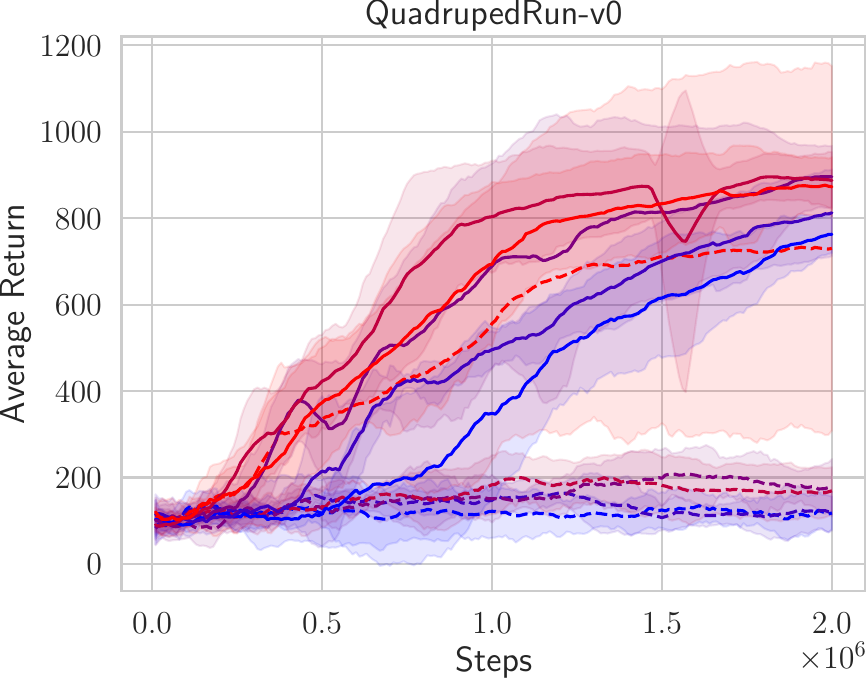} &
    \includegraphics[width=0.3\columnwidth]{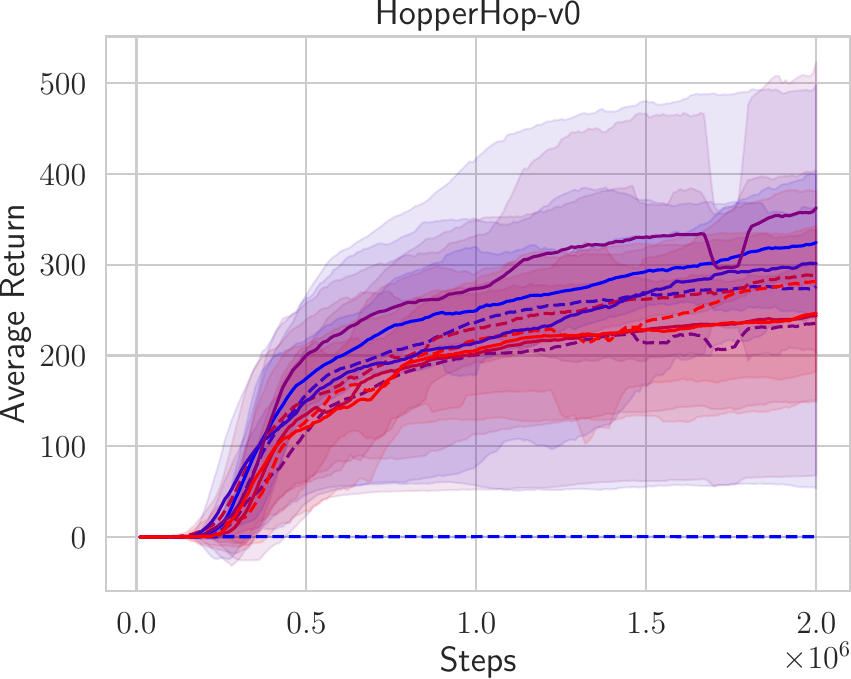} &
    \includegraphics[width=0.3\columnwidth]{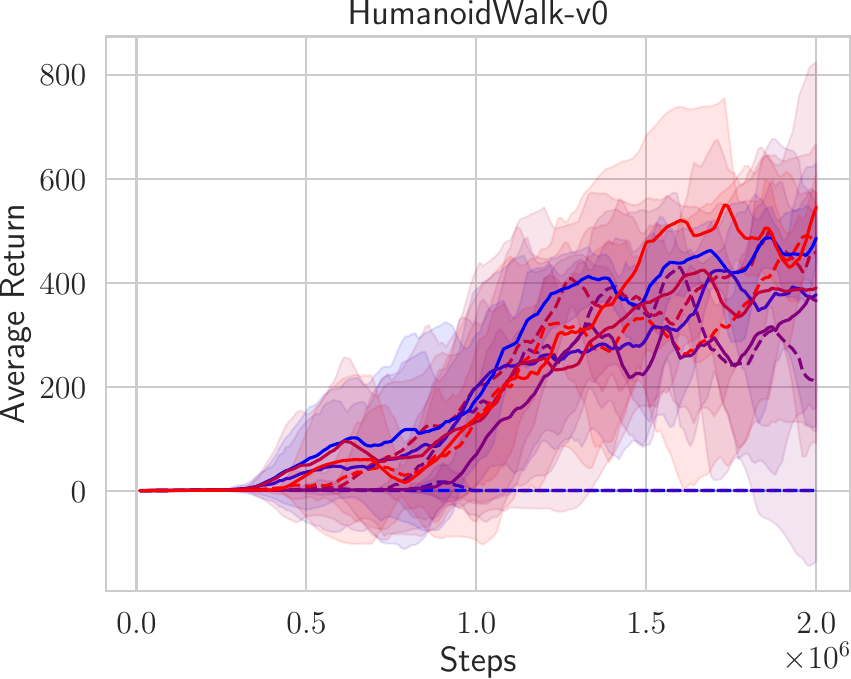} \\
    \includegraphics[width=0.3\columnwidth]{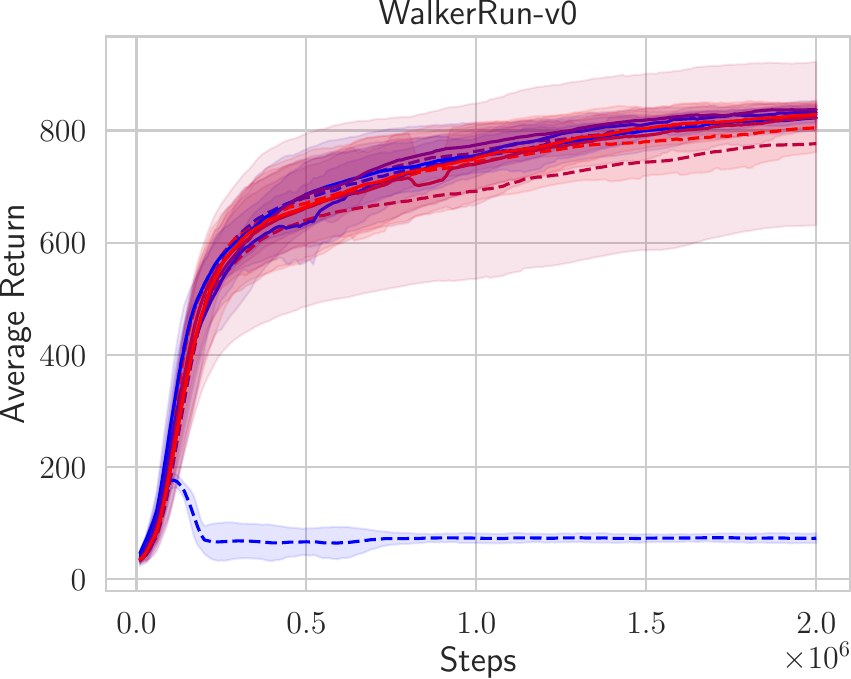} &
    \includegraphics[width=0.3\columnwidth]{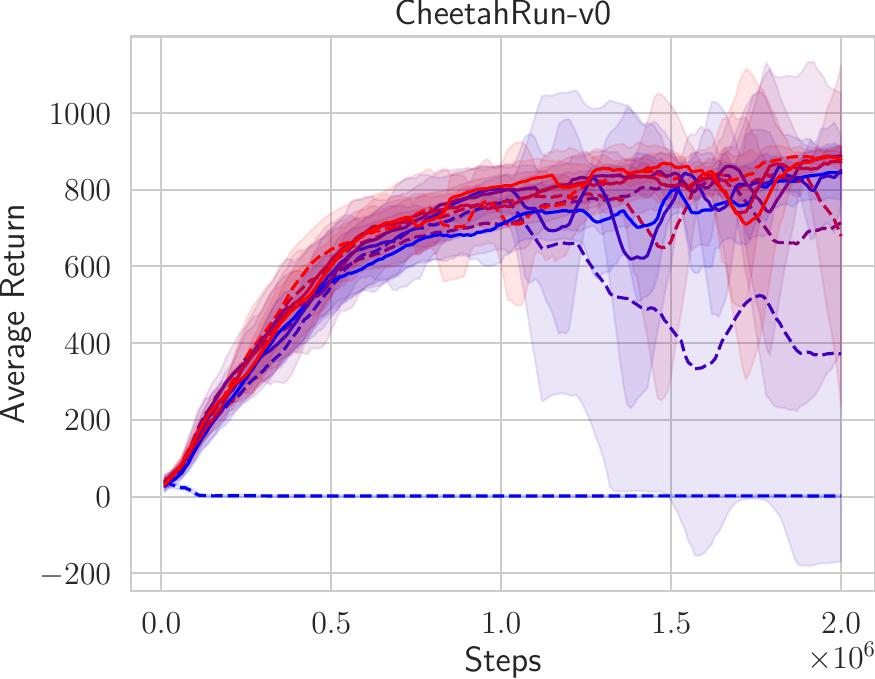} &
    \raisebox{0.7cm}{\includegraphics[width=0.3\columnwidth]{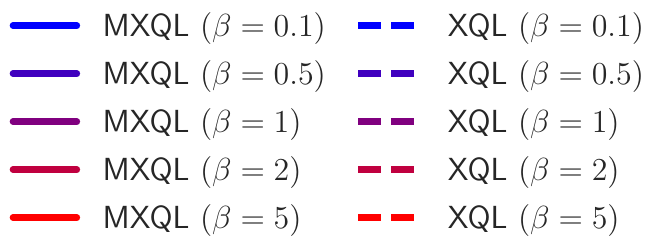}}
  \end{tabular}
  \caption{Performance comparison of MXQL (n=8) and XQL on DM Control tasks for online RL.}
  \label{fig:online_plot}
\end{figure}

In online RL, experiments were conducted using five tasks from the DM control suite.
These tasks include Quadruped-run, Hopper-hop, Walker-run, Cheetah-run, as experimented in \cite{garg2023extreme}, in addition to Humanoid-walk.
Both XQL and MXQL used SAC \citep{haarnoja2018soft} as the base algorithm.
Experiments were conducted with a range of $\beta$ values ([0.1, 0.5, 1, 2, 5]), which includes additional smaller values ([0.1, 0.5]) not used in the experiments by \cite{garg2023extreme} ([1, 2, 5]).
In \cite{garg2023extreme}, when XQL became unstable, it was restarted, but in this study, no restarts were performed for a fair comparison.
The plot of average returns when the expansion order is $8$ is shown in \cref{fig:online_plot}.

In most tasks, MXQL demonstrated superior performance compared to XQL.
Particularly in cases of small $\beta$, XQL often failed to learn adequately.
This implies that XQL could not learn with a small trust region, whereas MXQL was able to stabilize learning even for small $\beta$.
In other words, MXQL allows for a wider range of $\beta$ selections.
The scores averaged across all tasks are shown in \cref{tab:online_result}.
Experiments were conducted with $n=4, 8, 12$, and stable learning was achieved across all values.
The final average scores and standard deviations for each task when using the best $\beta$ in XQL and MXQL ($n=8$) are shown in \cref{tab:appendix_online_best}, demonstrating improved performance with the tuned $\beta$.



\subsection{Offline RL}
To ensure a fair comparison with XQL, experiments were conducted on the same Gym locomotion, AntMaze, and Franka Kitchen tasks as those reported in \cite{garg2023extreme}.
The scores are displayed in \cref{tab:offline_result} and are compared with those from \cite{NEURIPS2021onestep,fujimoto2021td3bc,NEURIPS2020cql,kostrikov2022iql, garg2023extreme}.
The proposed method, MXQL, demonstrated the best scores in several tasks compared to the scores of existing methods reported in \cite{kostrikov2022iql, sikchi2024dual}, and outperformed XQL in most tasks.
In the AntMaze medium and large tasks, MXQL underperforms compared to IQL, suggesting that the soft optimal value estimated in both XQL and MXQL may not be well-suited for these particular tasks.

\begin{table}[htbp]
    \begin{center}
        \begin{tabular}{l||rrrrrr|r}
         Dataset & BC & Onestep RL & TD3+BC & CQL & IQL & XQL & MXQL \\
         \hline 
        halfcheetah-med  & 42.6 & \textbf{48.4} & \textbf{48.3} & 44.0 & 47.4 & 47.4 & 46.5 $\pm$ 0.3    \\ 
        hopper-med  & 52.9 & 59.6 & 59.3 & 58.5 & 66.3 & \textbf{68.5} & \textbf{68.3} $\pm$ 7.3 \\ 
        walker2d-med  & 75.3 & 81.8 & \textbf{83.7} & 72.5 & 78.3 & 81.4 & 71.9 $\pm$ 3.9   \\ 
        halfcheetah-med-rep  & 36.6 & 38.1 & \textbf{44.6} & \textbf{45.5} & 44.2 & 44.1 & 44.1 $\pm$ 0.3  \\ 
        hopper-med-rep & 18.1 & \textbf{97.5} & 60.9 & 95.0 & 94.7 & 95.1 & \textbf{98.2} $\pm$ 3.2  \\ 
        walker2d-med-rep  & 26.0 & 49.5 & \textbf{81.8} & 77.2 & 73.9  & 58.0 & 57.9 $\pm$ 8.8  \\ 
        halfcheetah-med-exp  & 55.2 & \textbf{93.4} & 90.7 & 91.6 & 86.7 & 90.8 & 88.2 $\pm$ 4.4 \\ 
        hopper-med-exp  & 52.5 & 103.3 & 98.0 & \textbf{105.4} & 91.5 & 94.0 & \textbf{105.1} $\pm$ 10.1 \\ 
        walker2d-med-exp & 107.5 & \textbf{113.0} & 110.1 & 108.8 & 109.6 & 110.1 & 109.9 $\pm$ 0.1 \\ 
        \hline
        antmaze-umaze & 54.6 & 64.3 & 78.6 & 74.0 & 87.5 & 47.7 & \textbf{88.3} $\pm$ 2.1 \\ 
        antmaze-umaze-div & 45.6 & 60.7 & 71.4 & \textbf{84.0} & 62.2 & 51.7 & 53.2 $\pm$ 9.7  \\ 
        antmaze-med-play & 0.0 & 0.3 & 10.6 & 61.2 & \textbf{71.2} & 31.2 & 50.8 $\pm$ 2.7 \\ 
        antmaze-med-div & 0.0 & 0.0 & 3.0 & 53.7 & \textbf{70.0} & 0.0 & 52.2 $\pm$ 6.6 \\ 
        antmaze-large-play & 0.0 & 0.0 & 0.2 & 15.8 & 39.6 & 10.7 &  18.7 $\pm$ 5.9  \\ 
        antmaze-large-div & 0.0 & 0.0 & 0.0 & 14.9 & \textbf{47.5} & 31.3 & 14.3 $\pm$ 6.4 \\ 
        \hline
        kitchen-complete & \textbf{65.0} & - & - & 43.8 & 62.5 & 56.7 & \textbf{64.2} $\pm$ 10.3  \\ 
        kitchen-partial & 38.0 & - & - & \textbf{49.8} & 46.3 & 48.6 & 47.1 $\pm$ 8.7  \\ 
        kitchen-mixed & 51.5 & - & - & 51.0 & 51.0  & 40.4 & \textbf{71.9} $\pm$ 3.6  \\ 
        \hline
        \end{tabular}
    \end{center}
    \caption{Average normalized scores on Gym locomotion, AntMaze and Kitchen tasks. Highlighted results are within one performance point of those achieved by the best-performing algorithm, and the standard deviation across six seeds is displayed as $\pm$ for MXQL.}
    \label{tab:offline_result}
\end{table}

\begin{table}[t]
  \centering
  \begin{minipage}[t]{0.6\textwidth}
    \centering
        \begin{tabular}{l||rrrr}
         \multirow{2}{*}{$\beta$} & \multirow{2}{*}{XQL} & \multicolumn{3}{c}{MXQL} \\ 
                 & & n=4 & n=8 & n=12       \\ 
         \hline
         0.1 & 38.9 & \dashuline{619.0} & \dashuline{648.0} & \dashuline{642.9} \\
         0.5 & 319.2 & \dashuline{665.4} & \dashuline{634.8} & \dashuline{598.9} \\
         1 & 432.1 & \dashuline{653.9} & \dashuline{669.9} & \dashuline{622.8} \\
         2 & 474.1 & 626.5 & 643.3 & \dashuline{643.6}\\
         5 & 638.7 & 654.3 & 674.7 & 642.0\\
         \hline
        \end{tabular}
    \caption{Average scores of XQL and MXQL with various $n$ values for five tasks from DM Control in online RL. Scores underlined with dashes represent significant differences between XQL and MXQL, as determined by a t-test with a significance level of 0.05.}
    \label{tab:online_result}
    \end{minipage}
  \hfill
\end{table}


\section{Related Work}
XQL \citep{garg2023extreme} has evolved based on MaxEnt RL \citep{a2014maxentmdp,a2017softq,haarnoja2018soft} and allowed for the estimation of soft-values through Gumbel loss without access to entropy, similar to \cite{NIPS2015svg,a2017softq}.
XQL has been successfully combined with existing methods widely used in online RL, such as \cite{fujimoto2018td3,haarnoja2018soft}.
However, its instability and sensitivity to the value of $\beta$ necessitate careful tuning. In MXQL, the loss function has been stabilized through Maclaurin expansion. \cite{a2021logiql,hui2023doublegumbel} have also used loss functions different from L2.

In offline RL, some methods regularize through conservatism \citep{wu2019brac,NEURIPS2019bear,p2019bcq,NEURIPS2020cql,fujimoto2021td3bc,nair2021awac} and others that directly model the greedy policy \citep{peng2019advantageweighted,NEURIPS2021onestep,NEURIPS2021dt}.
XQL was able to learn without direct access to the current policy, similar to \cite{kostrikov2022iql, xu2023sqleql}, while introducing conservatism.
Similar to online RL, methods for stabilization have been employed in offline RL; notably, the stabilization in MXQL has shown performance improvements in certain offline RL tasks.
Recently, there have been methods that have led to performance improvements by using Diffusion Models \citep{pmlr2015diffusion1,NEURIPS2020diffusion2,song2021diffusion3} as the policy \citep{wang2023diffusionpolicies,hansenestruch2023idql}, and applying this approach in MXQL could be a future direction.

\section{Conclusion}
A recent study \citep{garg2023extreme} suggested that the Bellman error might follow a Gumbel distribution, challenging traditional least squares methods. 
In response, XQL, utilizing Gumbel Regression, was proposed but encountered stability issues in online RL due to extreme gradient values and the often incorrect assumption that state-action pairs are i.i.d..
The study introduces Maclaurin Expanded Extreme Q-learning (MXQL), stabilizing Gumbel loss and allowing adjustment between normal and Gumbel error distributions using Expanded Gumbel loss. This modification offers a practical solution to the instability and error distribution assumption issues seen in XQL, showing improved performance and stability in both online and offline RL scenarios.

\subsubsection*{Acknowledgments}

This work was partially supported by JST Moonshot R\&D Grant Number JPMJPS2011, CREST Grant Number JPMJCR2015 and Basic Research Grant (Super AI) of Institute for AI and Beyond of the University of Tokyo.
T.O. was partially supported by JSPS KAKENHI Grant Number JP23K18476.

\bibliography{main}
\bibliographystyle{rlc}

\appendix

\section{Experiments}
\subsection{Online RL}
\begin{figure}[h]
  \centering
  \setlength{\tabcolsep}{2pt}
  \begin{tabular}{lcr}
    \includegraphics[width=0.3\columnwidth]{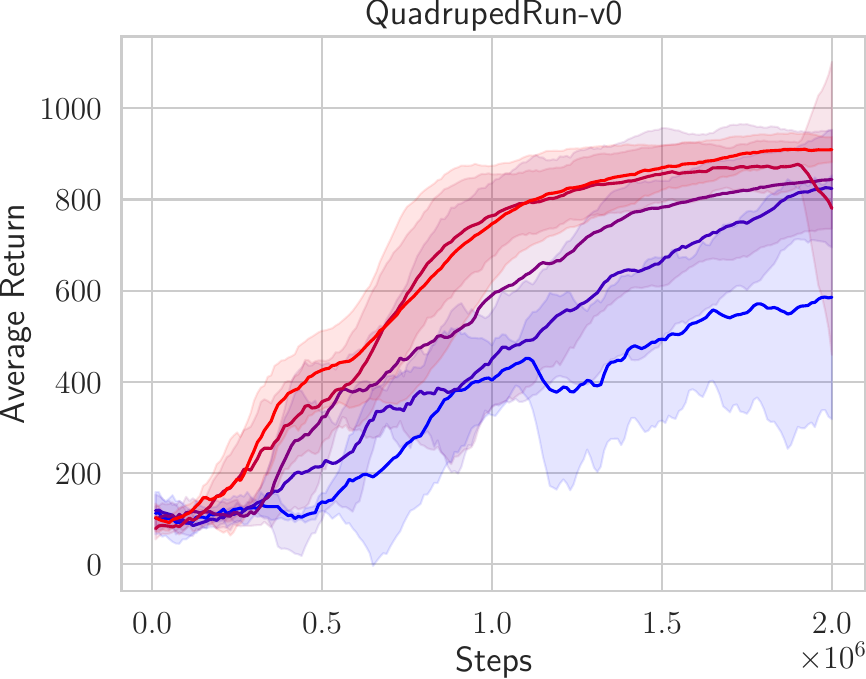} &
    \includegraphics[width=0.3\columnwidth]{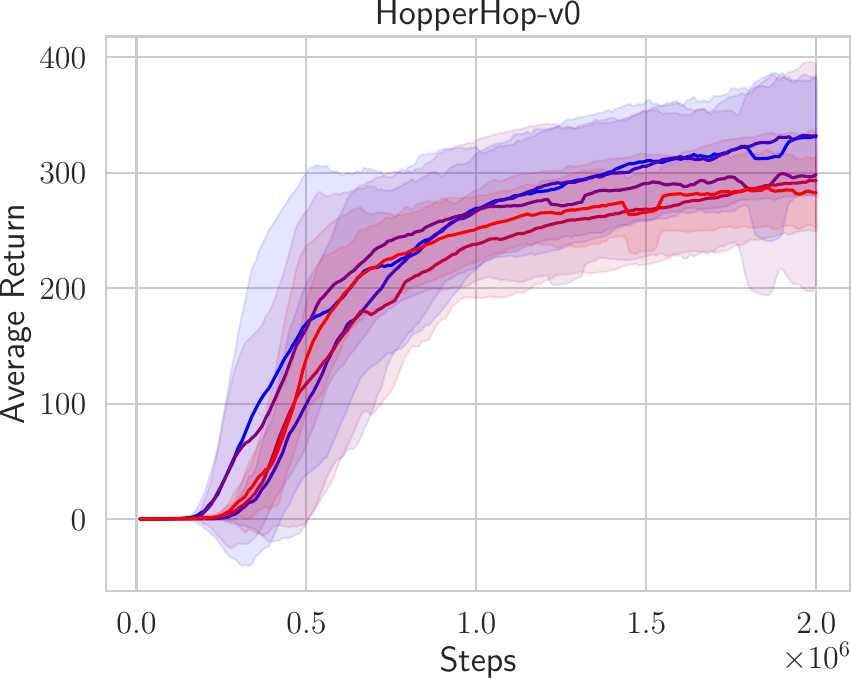} &
    \includegraphics[width=0.3\columnwidth]{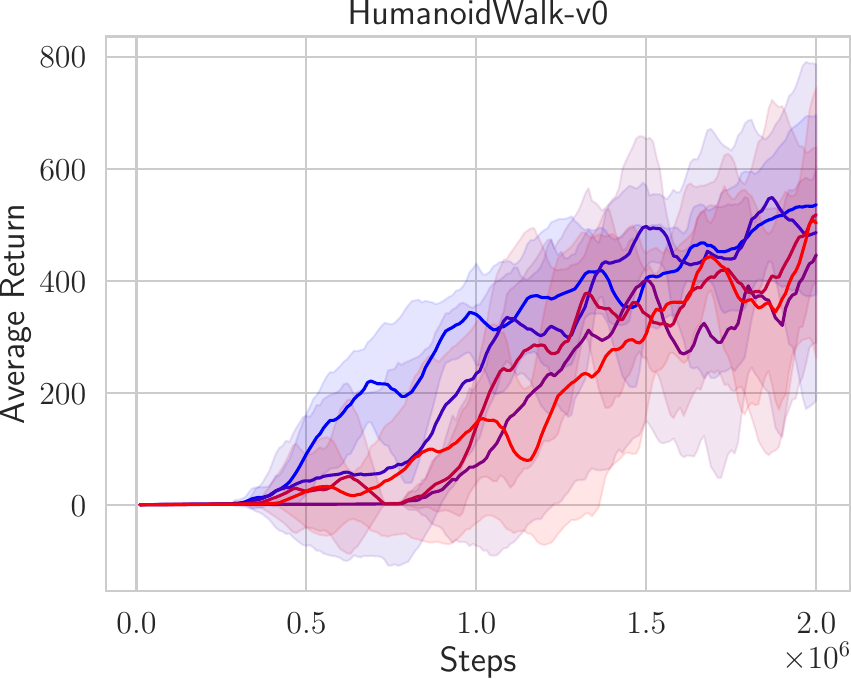} \\
    \includegraphics[width=0.3\columnwidth]{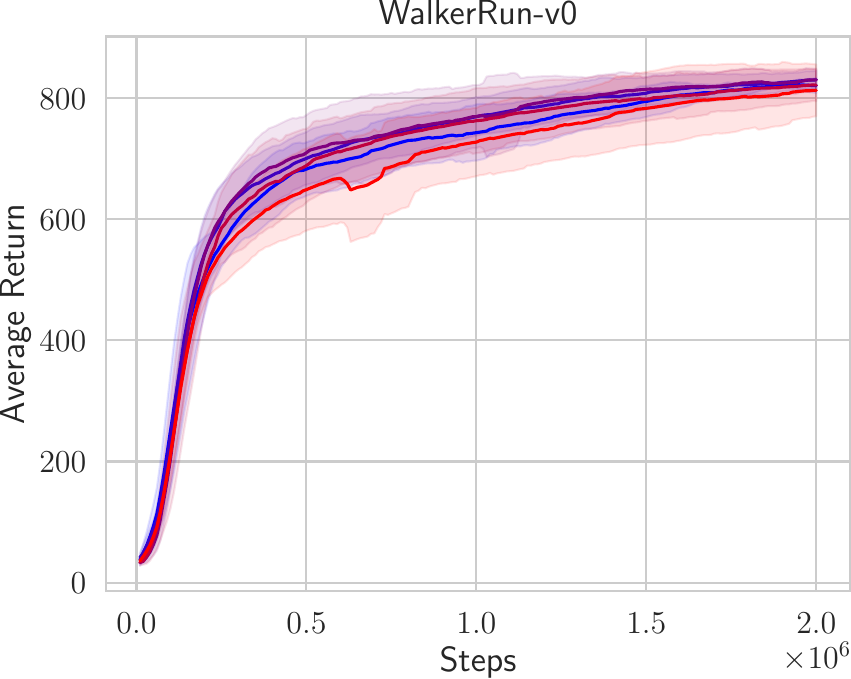} &
    \includegraphics[width=0.3\columnwidth]{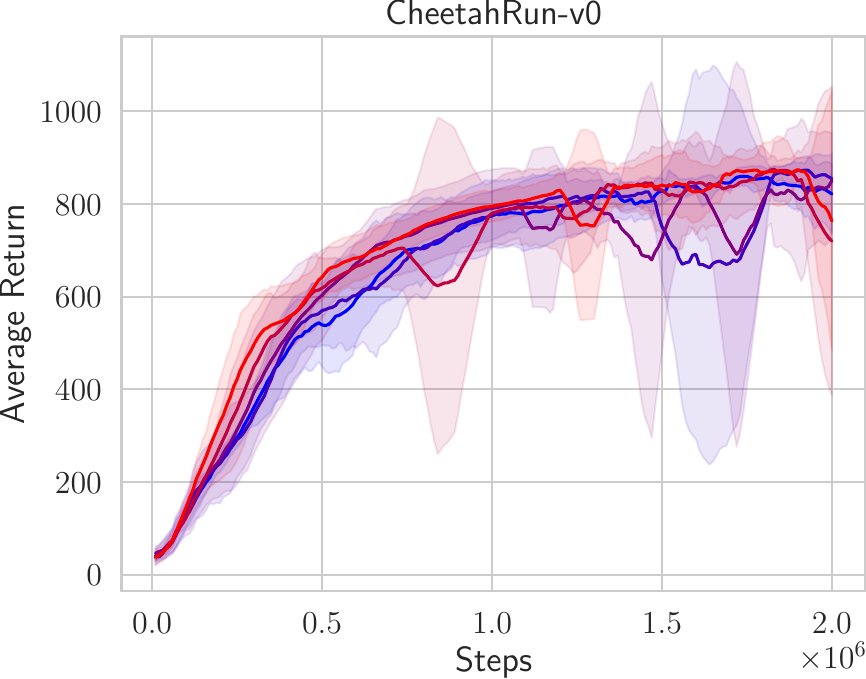} &
    \raisebox{0.cm}{\includegraphics[width=0.25\columnwidth]{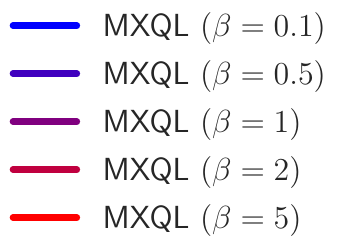}}
  \end{tabular}
  \caption{Performance of MXQL (n=4) in the online RL tasks from DM Control.}
  \label{fig:appendix_online_n4}
\end{figure}

\begin{figure}[h]
  \centering
  \setlength{\tabcolsep}{2pt}
  \begin{tabular}{lcr}
    \includegraphics[width=0.3\columnwidth]{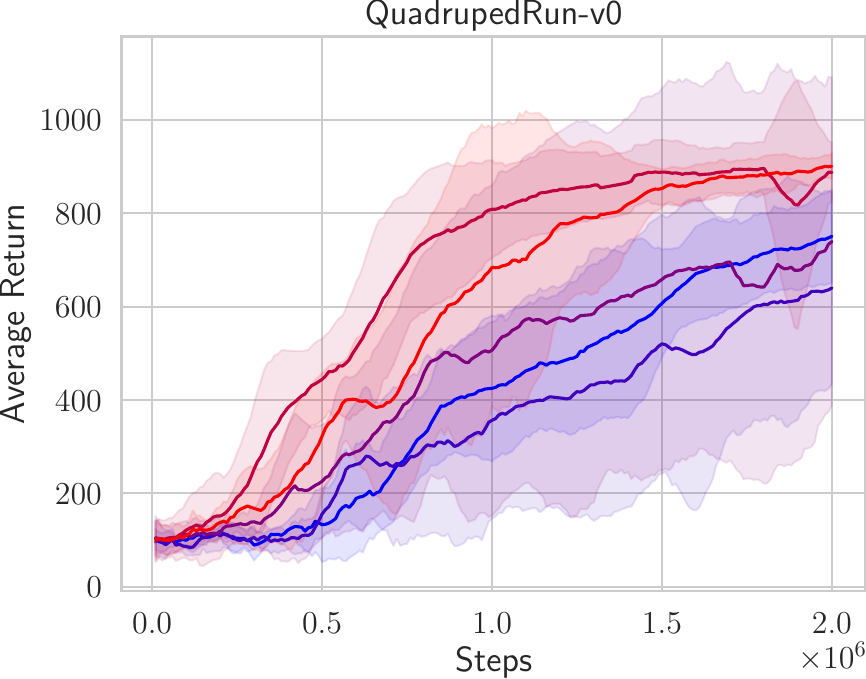} &
    \includegraphics[width=0.3\columnwidth]{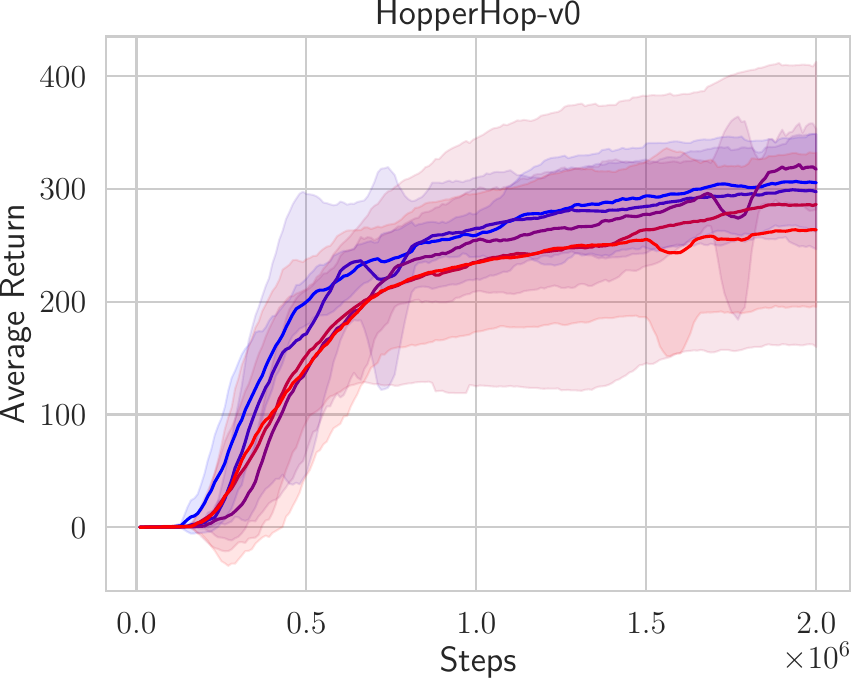} &
    \includegraphics[width=0.3\columnwidth]{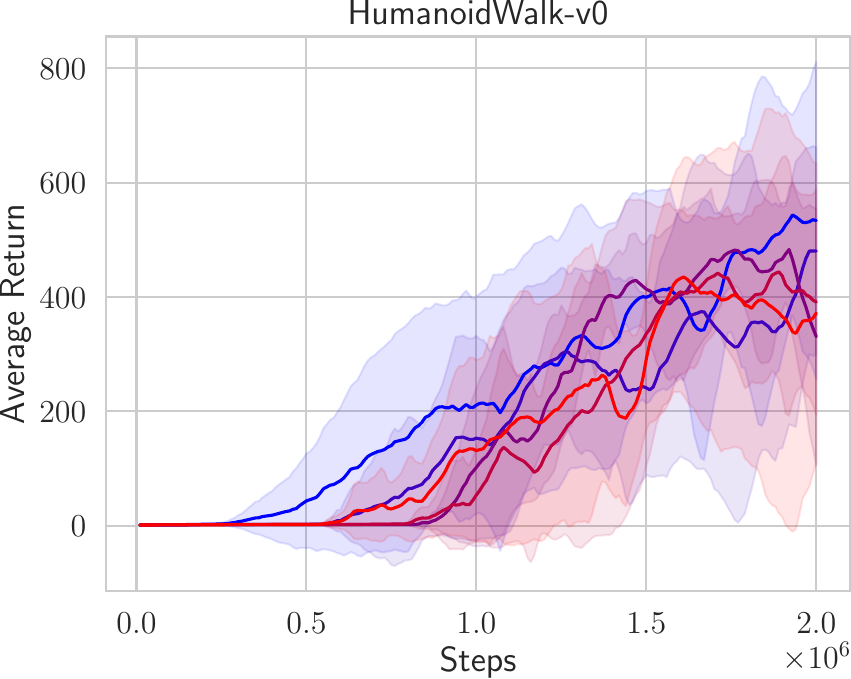} \\
    \includegraphics[width=0.3\columnwidth]{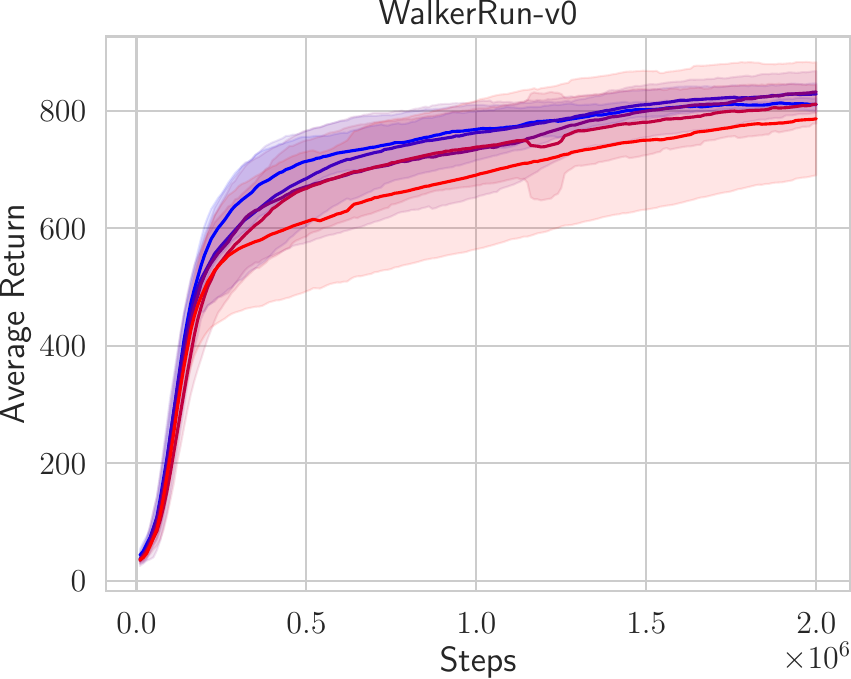} &
    \includegraphics[width=0.3\columnwidth]{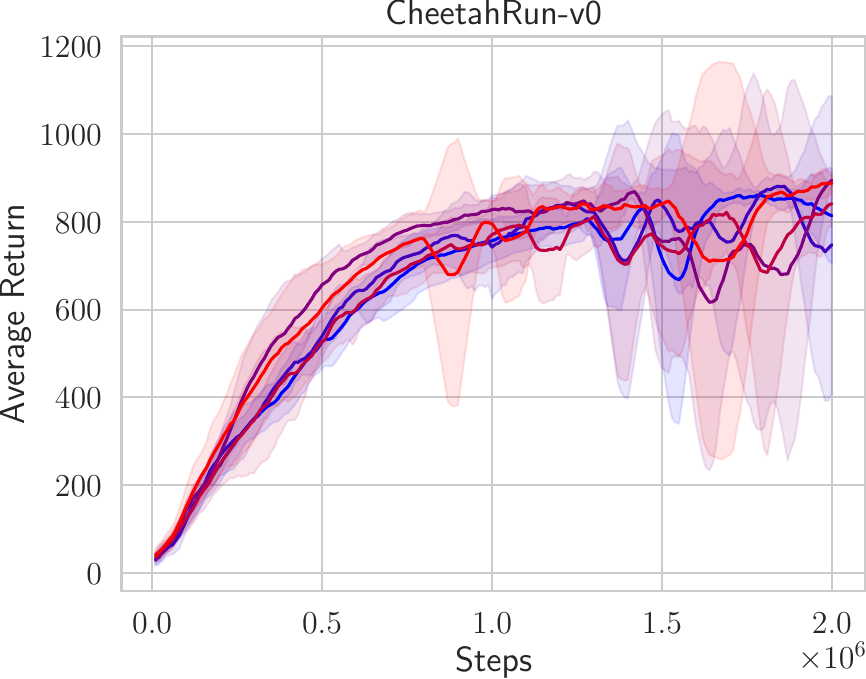} &
    \raisebox{0.cm}{\includegraphics[width=0.25\columnwidth]{figure/onlinerl/label_mxql_only.pdf}}
  \end{tabular}
  \caption{Performance of MXQL (n=12) in the online RL tasks from DM Control.}
  \label{fig:appendix_online_n12}
\end{figure}

In the experiments of MXQL and XQL, the implementation is based on the official implementation of \citep{garg2023extreme}, with only the loss function changed. The hyperparameters are the same as in \citep{garg2023extreme}. In all experiments on online RL, 5 seeds are used, and the 95\% confidence interval is shown as a shaded area. The results of changing the order $n$ of expansion in MXQL are shown in \cref{fig:appendix_online_n4} and \cref{fig:appendix_online_n12}.

\begin{table}[h]
    \centering
        \begin{tabular}{l||rr|rr} 
         \multirow{2}{*}{Task} & \multicolumn{2}{c}{XQL} & \multicolumn{2}{c}{MXQL} \\
         & $\beta$ & Score & $\beta$ & Score \\ 
         \hline
         QuadrupedRun-v0 & 5 & 730.2 $\pm$ 303.8 & 1 & 896.0 $\pm$ 51.4\\
         HopperHop-v0 & 2 & 287.4 $\pm$ 9.1 & 1 & 362.7 $\pm$ 115.7 \\
         HumanoidWalk-v0 & 5 & 487.1 $\pm$ 60.3 & 5 & 546.1 $\pm$ 45.0\\
         WalkerRun-v0 & 0.5 & 826.0 $\pm$ 19.4 & 1 & 837.2 $\pm$ 5.3\\
         CheetahRun-v0 & 5 & 890.0 $\pm$ 16.9 & 1 & 887.6 $\pm$ 7.6\\
        \hline
        \end{tabular}
    \caption{The final average score and standard deviation when using the best $\beta$ in XQL and MXQL (n=8).}
    \label{tab:appendix_online_best}
\end{table}

\subsection{Offline RL}

In offline RL, the official implementation is also used, and the hyperparameters are the same as in \cite{garg2023extreme}. In Gym tasks, "-v2" was used, while in AntMaze and Kitchen tasks, "-v0" was employed. The batch size and the update frequency of the V-function, which were tuned in \citep{garg2023extreme}, are not tuned. 
In XQL, the $\beta$ and the size of clipping, and in MXQL, the $\beta$ and the order $n$ of expansion have been tuned, and a common value for each domain has been used. The range for tuning $\beta$ is the same as for \cite{garg2023extreme}, which is  [0.6, 0.8, 1, 2, 5].
The $n$ was selected from $[4,8,12,16,20]$.
These values are shown in \cref{tab:appendix_offline_param}.
In \cite{garg2023extreme}, the evaluation was based on the best score during the learning process rather than the final score, and therefore, the XQL scores in \cref{tab:offline_result} are cited from \cite{sikchi2024dual}.
The experiments are conducted using 6 random seeds.

\begin{table}[h]
    \centering
        \begin{tabular}{l||rr|rr} 
         \multirow{2}{*}{Dataset} & \multicolumn{2}{c}{XQL} & \multicolumn{2}{c}{MXQL} \\
         & $\beta$ & Clip & $\beta$ & $n$ \\ 
         \hline
         Gym & 2 & 7 & 2 & 20 \\
         AntMaze & 0.6 & 7 & 1 & 8 \\
         Kitchen & 5 & 7 & 1 & 4 \\
        \hline
        \end{tabular}
    \caption{Hyperparameters in offline RL tasks from D4RL. The hyperparameters for XQL are the same as those used in \citep{garg2023extreme}.}
    \label{tab:appendix_offline_param}
\end{table}

\section{Details of the Figures}

In the calculation of the error distribution in the right of \cref{fig:egloss}, the loss function, which is the log-likelihood, is applied with "-exp", and a coefficient for normalization is multiplied. This coefficient is calculated to ensure that the integral of the distribution equals one, using "scipy.integrate.quad" \citep{2020SciPy-NMeth}.

In the preliminary experiments of Gumbel Regression in \cref{fig:regression}, the estimation was performed using stochastic gradient descent with 10,000 data. The learning rate was set at 0.02, and the batch size was 32. The mean and standard deviation were calculated across 100 experiments.



\end{document}